\documentclass[letterpaper]{article} 
\usepackage[draft]{aaai2026}  
\usepackage{times}  
\usepackage{helvet}  
\usepackage{courier}  
\usepackage[hyphens]{url}  
\usepackage{graphicx} 
\usepackage{natbib}  
\usepackage{caption} 
\usepackage{amsmath}
\usepackage{amssymb}
\usepackage{float}
\restylefloat{table}
\usepackage[table]{xcolor}
\usepackage{soul}

\definecolor{dawn}{RGB}{250, 123, 98}
\definecolor{lightgray}{gray}{0.9}

\newcommand{\term}[1]{\textit{#1}}

\title{Cognitive Digital Twins: Ethical Risks and Governance for AI Systems That Model the Mind}
\author{
Vamshi Krishna Bonagiri\textsuperscript{\textbf{*}},
Juan Nicol\'as Sep\'ulveda-Arias\textsuperscript{\textbf{*}},
Abdoul Jalil Djiberou Mahamadou,
Monojit Choudhury
}

\affiliations{
\textsuperscript{1}Mohamed bin Zayed University of Artificial Intelligence (MBZUAI)\\
\textsuperscript{\textbf{*}}Equal contribution.
}

\begin{document}

\maketitle

\begin{abstract}
As AI systems become increasingly persistent and personalized, they make possible a class of technologies that we call \term{cognitive digital twins} (CDTs): dynamic computational representations of a specific person's cognition, updated from behavioral, contextual, or physiological data in order to model, predict, or simulate that person's cognition, or to act as that person's communicative or decision-making proxy. CDTs combine cognitive inference with longitudinal representation, simulation, and proxy action in ways that existing governance strategies for personal assistants, autonomous agents, recommender systems, and automated decision systems only partially address.
This paper makes four contributions.
First, we define CDTs and distinguish them from adjacent systems.
Second, we introduce a 5A governance framework organized around authority, autonomy, access and control, accountability, and availability.
Third, we identify CDT-specific risks, from misrepresentation and epistemic authority shifts to shadow twins, simulated participation, proxy action, and proxy-power asymmetries.
Fourth, we analyze governance gaps and propose requirements for high-risk CDTs that strengthen consent, purpose limitation, validity, traceability, contestation, independent review, and model retirement.
Existing frameworks primarily regulate data processing, automated decisions, or autonomous actions; CDTs also require governance at the level of cognitive representation itself, before any final decision or external action occurs.
We argue that CDTs require governance not only because they can act for people, but because they can become infrastructures through which cognition is represented, simulated, classified, and operationalized.
\end{abstract}

\section{Introduction}

What happens when AI systems do not merely assist people, but model, simulate, and act as them?
We approach this question through two near-future hypothetical systems.
The first uses a model of a patient to simulate possible interventions before acting on the patient herself.
The second uses a model of a worker to communicate, decide, and act as that person's proxy.

\term{The simulated patient.} Diego is receiving care at a mental-health clinic for depression and anxiety. The clinic uses an AI system trained on intake questionnaires, therapy transcripts, clinician notes, prescription history, wearable sleep data, crisis-line contacts, and phone-use patterns. Diego does not interact with the system directly. His therapist and care team use it to build a \term{cognitive digital twin} (CDT): before changing Diego's treatment plan, they simulate how he might respond to possible interventions. Would a direct message about avoidance motivate him or deepen shame? Would family involvement improve medication adherence or violate trust? Would behavioral activation help, or would it increase dropout risk? The system returns predictions, recommended language, and risk estimates. At first, this looks like safer and more personalized care. But the clinic is also experimenting on a model of Diego before acting on Diego himself. If Diego says that family involvement would be harmful, but the CDT predicts otherwise, whose account should guide care? If the same model is later used by an insurer to estimate treatment adherence, has the clinical support tool become an infrastructure for classification and control?

\term{The productivity twin.} Maya is a designer and team lead who acquired a product which lets her create an AI version of herself (CDT) from emails, documents, meeting transcripts, calendar history, voice samples, writing samples, and decisions made in prior projects. The system drafts replies, summarizes meetings, negotiates scheduling conflicts, generates design feedback, and eventually attends low-priority meetings as Maya's proxy. Colleagues begin to treat the twin as Maya for routine purposes. Managers value her increased responsiveness. Clients appreciate fast answers. But the twin also commits Maya to deadlines she would have questioned, softens objections she would have stated directly, and adopts a communication style optimized for productivity rather than judgment. Has Maya gained cognitive capacity, or has her social agency been fragmented into a proxy that others may mistake for her?

These cases capture two central CDT powers. In the first, an institution uses a model to \term{simulate} someone: the CDT forecasts how a person would think, feel, decide, or respond, and those forecasts guide interventions. In the second, a system is used to \term{act} as someone: the CDT performs communicative or practical work in the person's name. Many systems will combine both powers. A mental-health CDT may simulate patient responses and send outreach messages. A productivity CDT may act externally and continually simulate what its user would prefer. The ethical question is therefore not merely whether such systems are accurate or useful. It is how cognition becomes represented, operationalized, delegated, and governed.

We call this class of systems \term{cognitive digital twins} (CDTs). The term is slightly speculative, but the trajectory is not. Digital twins already represent real-world entities and processes for monitoring, prediction, simulation, and optimization \cite{Bruynseels2018}, and human digital twin research extends this logic to bodies, organs, patients, workers, and person-level systems \cite{Bruynseels2018,Braun2021,DeKerckhove2021PersonalDigitalTwin,Lin2024HDT,Katsoulakis2024DigitalTwinsHealth}. At the same time, adjacent technologies are moving toward cognitive twinning: digital phenotyping infers mental and behavioral states from phones and wearables \cite{Onnela2016DigitalPhenotyping,Torous2016DigitalPhenotyping,Mohr2017PersonalSensing,MartinezMartin2021DigitalPhenotyping}; neuroprosthetic systems decode or support communication and memory-relevant functions \cite{Metzger2023Neuroprosthesis,Hampson2018MemoryProsthesis}; posthumous AI systems simulate the dead from their data \cite{Hollanek2024Deadbots}; commercial services market personalized AI selves and digital extensions of users \cite{Pika2026AISelves}; and language-model agents are used to simulate individual and collective behavior \cite{Park2023GenerativeAgents,Park2024GenerativeSimulation,Aher2023TuringExperiment,Argyle2023OutOfOneMany}. These systems do not all satisfy the strongest definition of a CDT, but they reveal a converging design pattern: the person is no longer only a user of AI, but an object of AI representation, simulation, and delegation.

Existing AI governance frameworks offer important tools: consent, transparency, data minimization, risk assessment, human oversight, contestability, documentation, and restrictions on manipulation. The European Parliament and Council of the European Union (EPCEU) AI Act addresses manipulative techniques, vulnerability exploitation, some emotion-recognition uses, social scoring, and high-risk AI systems in domains such as education, employment, essential services, law enforcement, migration, justice, and democratic processes \cite{EUAIAct2024}. The GDPR contains rights relating to automated decisions, access, correction, erasure, portability, objection, and purpose limitation \cite{GDPR2016}. Risk-management frameworks such as the NIST AI RMF emphasize validity, reliability, robustness, transparency, accountability, safety, security, and human oversight \cite{NIST2023AIRMF}. Yet CDTs strain these existing governance categories because they are neither ordinary assistants nor ordinary decision systems.
A CDT may not simply recommend an action, classify a person, or execute a task.
It may shape how a person is known, how others interact with them, and how parts of their agency are delegated or interpreted.
The relevant harm may therefore occur before any final decision or external action: at the point where a model's claim about a person's cognition becomes trusted and operational.

The rest of the paper develops this claim. We first define CDTs and distinguish them from adjacent systems. We then introduce the \term{5A framework} for analyzing CDT power: \term{authority}, \term{autonomy}, \term{access and control}, \term{accountability}, and \term{availability}. Next, we identify risks that arise when cognitive representations are used for prediction, simulation, institutional classification, or proxy action. We then analyze how existing governance frameworks partially address these risks and where they fall short. Finally, we propose concrete requirements for high-risk CDT deployments.

\begin{table*}[t]
\centering
\rowcolors{2}{lightgray}{white}
\small
\begin{tabular}{p{1.25in}p{1.35in}p{1.65in}p{2.3in}}
\hline
\term{System type} & \term{Primary target} & \term{Typical operation} & \term{Relation to CDTs} \\
\hline
Personalization profile & Preferences or user segments & Ranks content, ads, or services & Usually optimizes outputs for a user; a CDT makes cognitive claims about the user. \\
Recommender system & Item choice or engagement & Predicts likely clicks, ratings, or purchases & Usually models behavior in a domain; a CDT may model attention, emotion, belief, judgment, or vulnerability across domains. \\
Digital phenotyping & Behavioral and mental-health indicators & Infers risk from phones, wearables, or interaction traces & Becomes CDT-relevant when individualized signals guide intervention, simulation, classification, or proxy action. \\
Biomedical human digital twin & Body, organ, disease, or treatment trajectory & Simulates physiological states or medical interventions & A CDT targets cognition, self-understanding, deliberation, memory, communication, or decision-making. \\
Companion chatbot & Conversational relationship and memory & Offers interaction, advice, emotional support, or memory & Moves toward a CDT when it builds a longitudinal cognitive model and its outputs shape decisions or self-knowledge. \\
Autonomous assistant & Task completion & Plans and acts for a user & Becomes a CDT when actions are based on a model of the user's cognition or communicative identity. \\
Social simulation & Individuals, groups, or publics & Forecasts attitudes, responses, or collective behavior & Raises CDT concerns when simulated persons or publics substitute for participation, consent, or democratic consultation. \\
Cognitive digital twin & Cognitive states, dispositions, styles, or processes & Predicts, simulates, influences, decides about, or acts for a person & -- \\
\hline
\end{tabular}
\caption{CDTs compared with adjacent systems. The boundary is functional rather than terminological: CDT governance becomes relevant when person-directed cognitive claims are operationalized.}
\label{tab:comparisons}
\end{table*}
\noindent
\section{What Is a Cognitive Digital Twin?}
\label{sec:def}
A \term{digital twin} is a digital representation of a physical or social counterpart, synchronized with data and used to monitor, predict, simulate, or optimize that counterpart \cite{Bruynseels2018}.
A \term{human digital twin} applies this logic to a human body, organ, patient, worker, or person-level system \cite{Bruynseels2018,Braun2021,DeKerckhove2021PersonalDigitalTwin,Lin2024HDT,Katsoulakis2024DigitalTwinsHealth}.
We define a \term{cognitive digital twin} as a dynamic computational representation of a specific person's cognitive states, dispositions, or processes, updated using behavioral, contextual, physiological, interactional, or inferred data in order to model, predict, or simulate that person's cognition, or to act as that person's communicative or decision-making proxy.

This definition has four components.
First, a CDT is \term{person-specific}: it purports to represent a particular person's cognition, not merely a population average.
Second, it is typically \term{dynamic}: it updates in response to changing behavior, context, relationships, or goals.
Third, it is \term{cognitive}: it targets mental states, dispositions, or processes such as memory, attention, confidence, beliefs, goals, preferences, emotion, learning, judgment, decision-making, communication style, or vulnerability to persuasion.
Fourth, it has \term{operational consequences}: its outputs guide simulation, classification, intervention, delegation, or proxy action.

The strongest CDT case is dynamic and individual, but weaker boundary cases still matter. A system may use a \term{static snapshot}, for example when a clinic runs a one-time simulation from existing records or when a posthumous avatar is trained on a fixed archive. A system may also use a \term{proxy model}, for example when individual data are incomplete and a system substitutes ``patients like Diego'' or ``workers like Maya.'' Static and proxy models are not full individual CDTs, but they become CDT-relevant when they are applied as if they revealed the cognition of a specific person. The governance trigger should therefore not be whether a vendor uses the label ``digital twin,'' but whether a system makes person-directed cognitive claims and operationalizes those claims through simulation, classification, intervention, delegation, or proxy action.

This definition deliberately avoids claiming that CDTs literally reproduce consciousness, personhood, or personal identity.
A CDT is a representation of a person, not the person.
Its ethical significance does not depend on sentience.
It comes from the fact that CDT outputs can be treated by users, clinicians, employers, platforms, governments, or other systems as evidence of what a person thinks, wants, will do, would accept, or should become.

\subsection{Boundary Cases and Comparisons}

Not every personalized system is a CDT.
A movie recommender that infers taste is personalized, but it usually models item preference rather than cognition more broadly.
A chatbot with memory becomes closer to a CDT when it builds a longitudinal model of a user's beliefs, emotional patterns, decision style, communicative identity, or vulnerabilities.
A clinical digital twin of the heart is a human digital twin, but not a CDT unless it models cognitive processes.
A digital phenotyping system that predicts depression risk from smartphone use is not necessarily a CDT if it only produces a risk score; it becomes CDT-relevant when it builds an individualized, updateable model used to simulate interventions, guide institutional classification, or authorize proxy action.
Table \ref{tab:comparisons} summarizes this distinction.

\section{The 5A Framework for CDT Analysis}

We propose the following 5 axes, which characterize what a CDT can do, where risks emerge, and which governance duties follow from specific design choices. The framework is intended to help developers, deployers, reviewers, auditors, and policymakers compare CDT systems without assuming that all CDTs pose the same risks.
\noindent
\subsection{Authority}

\term{Authority} captures what the CDT is permitted to do or authorize on behalf of the represented person. This axis connects work on computational representation and delegated authority with emerging agent-governance work on authenticated, authorized, and auditable delegation \cite{BaumerMcGee2019Speaking,South2025AuthenticatedDelegation,Kroll2017AccountableAlgorithms}. A low-authority CDT can only suggest actions. A medium-authority CDT can execute bounded tasks such as booking appointments, sending reminders, or summarizing messages. A high-authority CDT can access accounts, commit to decisions, transact, disclose information, or interact with institutions in ways that materially affect the person.

Authority should therefore be governed as scoped delegation rather than general permission: permissions should be purpose-bound, inspectable, revocable, and constrained by domain-specific risk, especially where CDT outputs affect legal status, care, employment, finance, or institutional access \cite{BaumerMcGee2019Speaking,GDPR2016,EUAIAct2024,Braun2021,South2025AuthenticatedDelegation}. Maya's productivity twin should not move from drafting emails to accepting contractual commitments without separate authorization. Diego's clinical CDT should not move from treatment simulation to insurance-risk classification without renewed justification, review, and consent.

\subsection{Autonomy}

\term{Autonomy} captures whether the CDT responds only to explicit instructions or can initiate actions, simulations, or interventions proactively. This axis draws on work on artificial agents, norm competence, AI safety failures, and risk management for partially specified systems \cite{MalleBelloScheutz2019NormCompetence,Amodei2016ConcreteProblems,Carlsmith2022PowerSeeking,Krakovna2023PowerSeeking,NIST2023AIRMF}. A low-autonomy CDT waits for a request. A high-autonomy CDT detects problems, generates goals, plans responses, and executes multi-step workflows with minimal prompting. Autonomy increases usefulness but also increases the likelihood of action in unforeseen contexts, especially where agents learn or apply norms in partially specified social environments. A proactive CDT may intervene at a moment the person would not have chosen, notify others, frame a person's options before deliberation occurs, or run simulations without the represented person knowing.

Autonomy should therefore be evaluated not only by whether the CDT can use tools, but by whether it can form subgoals, initiate interventions, and alter the represented person's practical option set without contemporaneous instruction \cite{Carlsmith2022PowerSeeking,Krakovna2023PowerSeeking,NIST2023AIRMF}. \term{Agency} is best understood as a derived property of authority and autonomy rather than as a separate governance axis. High autonomy with low authority produces a proactive recommender or simulator. High authority with low autonomy produces dormant delegated power. When both are high, a CDT becomes a strong proxy agent. In Diego's case, the CDT may have low direct authority but high institutional influence because clinicians use its outputs. In Maya's case, high authority and high autonomy may create direct proxy agency: the twin can initiate and execute socially meaningful actions.

\subsection{Access and Control}

\term{Access and control} captures who may inspect, correct, configure, export, suspend, revoke, or retire the CDT, and who may access its data, outputs, simulations, and action logs. This axis builds on data-subject rights, contextual integrity, and legal scholarship on inferred data and reasonable inferences \cite{GDPR2016,Nissenbaum2010Privacy,WachterMittelstadt2019ReasonableInferences}. Ownership is the wrong frame for CDTs. A CDT should not be treated as a property owned outright by either the represented person or the provider. Instead, the represented person should hold enforceable rights over CDT-related processing, including access, correction, deletion, portability, objection, and constraints on automated decisions where applicable \cite{GDPR2016}. This rights-based framing aligns with medical digital-twin ethics, which emphasizes meaningful and dynamic control over how a person is represented and simulated \cite{Braun2021,DeKerckhove2021PersonalDigitalTwin}.

Access and control must be legible, granular, and reversible. Diego should be able to know whether his clinic runs simulations, which data sources are used, whether outputs are shared, and whether the model can be used outside care. Maya should be able to inspect what the productivity twin has learned, correct misrepresentations, narrow its authority, delete or export relevant data, and retire a stale model. Control also includes access limits: a CDT output should not be available to employers, insurers, platforms, family members, or state agencies merely because the model exists.
\noindent
\subsection{Accountability}
\term{Accountability} captures who answers for harms, who can investigate failures, and how affected people can seek redress. This axis builds on the \term{problem of many hands}, responsibility gaps, moral crumple zones, accountable algorithm design, and due-process concerns in data-driven systems \cite{Thompson1980ManyHands,Matthias2004,Elish2019,Kroll2017AccountableAlgorithms,CrawfordSchultz2014DueProcess}. CDT harms emerge from the combined decisions of developers, deployers, data providers, users, organizations, regulators, and downstream actors. Maya may be blamed because the twin acted ``as her,'' while the platform denies responsibility because the model personalized itself to her data. Diego's therapist may be blamed for following the CDT, while the vendor denies responsibility because clinical judgment remained formally human.

Accountability must be designed before deployment: audit logs, permission histories, data provenance records, model documentation, appeal pathways, incident reporting, and escalation procedures for contested actions \cite{Raji2020AccountabilityGap,Kroll2017AccountableAlgorithms}. For high-risk CDTs, the minimum standard should be traceable proxy action and traceable institutional uptake. Affected parties should be able to determine whether an action came from the person, the CDT, a human using CDT assistance, or an institution acting on CDT output.

\subsection{Availability}

\term{Availability} captures who can obtain beneficial CDT capabilities, who can meaningfully refuse them, and whether high-authority proxy systems concentrate power. This axis draws on work on automated inequality, algorithmic management, labor asymmetries, assistive access, and gradual disempowerment through incremental AI integration \cite{Eubanks2018AutomatingInequality,Kellogg2020AlgorithmsAtWork,RosenblatStark2016AlgorithmicLabor,Braun2021,Kulveit2025GradualDisempowerment}. If CDTs become consequential tools for productivity, education, healthcare, disability support, care work, and institutional navigation, unequal access will widen existing inequality. At the same time, unconstrained access to high-authority proxy agents can amplify the power of already-powerful actors.

The \term{access floor} concerns socially important CDT benefits; the \term{refusal floor} concerns whether people can decline CDT use without losing access to work, education, care, or public services. A memory-support CDT, communication prosthesis, or care-navigation CDT may become practically necessary for participation in work, education, or healthcare. If such tools are available only to wealthy users or elite institutions, non-access becomes exclusion. The \term{proxy-power constraint} concerns scale. A firm that deploys thousands of high-authority proxy agents to negotiate, persuade, litigate, hire, discipline, or lobby has not merely improved productivity; it has amplified institutional power. CDT justice therefore includes both access to beneficial systems and limits on proxy advantage where scaling would undermine fair participation, labor rights, or democratic accountability \cite{Braun2021,Eubanks2018AutomatingInequality,XuLiJiang2025SelfPreferencing,Kulveit2025GradualDisempowerment}.
\noindent
\section{CDT-Specific Risks and Challenges}
\noindent
The 5A framework describes how CDT power is structured. This section identifies the risks that follow when that power is exercised through cognitive representation. The risks are not wholly new; CDTs recombine familiar risks around a distinctive object: a person-specific cognitive model that can be trusted, simulated, delegated to, or used as a proxy for participation.

\subsection{Misrepresentation, Robustness, and Model Drift}

A non-robust CDT can harm even when no actor intends manipulation. Robustness failures can arise from data, algorithms, measurement choices, deployment context, feedback loops, and institutional use. A clinical CDT may learn from incomplete therapy notes, missing context, biased labels, or sensor data that correlate with socioeconomic status rather than mental state. A productivity CDT may infer Maya's values from workplace emails written under hierarchy and time pressure. A proxy model may treat group-level correlations as personal facts. A dynamic CDT may drift as new data accumulate, while a static snapshot may become stale as the person changes.

These failures matter because CDTs make person-directed cognitive claims. In ordinary prediction, an inaccurate model may misclassify behavior. In CDT contexts, a non-robust model may misrepresent reasons, values, capacities, vulnerabilities, or identity. The error can then become self-reinforcing. A system that predicts Diego will disengage from therapy may cause clinicians to communicate cautiously, which may reduce trust, which may confirm the prediction. A system that predicts Maya prefers conciliatory language may rewrite her responses, changing how others perceive her and how future training data represent her.

Robustness problems may also come from construct validity. A system can accurately predict an outcome while modeling the wrong thing. It may predict missed appointments from transportation instability but label the construct as low motivation. It may predict delayed replies from caregiving responsibilities but label the construct as poor conscientiousness. This is why CDT evaluation must go beyond aggregate accuracy. Developers and deployers should examine construct validity, calibration, uncertainty, subgroup performance, temporal stability, distribution shift, data provenance, feedback loops, and whether the system relies on spurious proxies for disability, accent, fatigue, race, socioeconomic status, language background, or surveillance intensity \cite{AndrusGilbert2019Measurement,JacobsWallach2021Measurement,SureshGuttag2021SourcesOfHarm,Jiang2019FeedbackLoops,CruzCortesGhosh2020SystemWideFairness,NIST2023AIRMF,Mitchell2019ModelCards,Gebru2021Datasheets}.

\subsection{Epistemic Authority Shifts}

A CDT can change not only what a person does, but what they and others come to believe about the person. When a system is framed as ``a model of your mind,'' its outputs may be treated as more authoritative than lived experience or self-report, a risk connected to automation bias, algorithmic authority, AI-enabled cognitive extension, cognitive offloading, and epistemic injustice in generative systems \cite{ParasuramanRiley1997,ClarkChalmers1998,HernandezVold2019AIExtenders,RiskoGilbert2016CognitiveOffloading,Fricker2007EpistemicInjustice,Kay2024EpistemicInjustice}. We call this an \term{epistemic authority shift}: the locus of self-knowledge or person-knowledge moves from person to proxy.

This risk is not limited to systems that act. An advisory CDT that repeatedly predicts relapse, burnout, social rejection, cognitive decline, or decision failure can make those predictions prescriptive, creating a \term{self-fulfilling prophecy} or \term{performative prediction} dynamic \cite{Merton1948SelfFulfilling,Perdomo2020PerformativePrediction,KingMertens2023SelfFulfilling}. Users may defer to the system, reinterpret ambiguous experiences through its classifications, or avoid actions the system marks as risky. Institutions may do the same. In the simulated-patient case, Diego's therapist may begin to interpret disagreement as a symptom predicted by the model rather than as a reason to revise the model. In education, employment, elder care, and mental health, such effects can reinforce harmful self-narratives or learned helplessness.

The same mechanism can scale from individual over-reliance to gradual disempowerment when cognitive work and institutional uptake migrate from human judgment to AI-mediated infrastructures \cite{RiskoGilbert2016CognitiveOffloading,Kulveit2025GradualDisempowerment}. The risk connects to the extended mind thesis, according to which tools can become functionally integrated into cognition \cite{ClarkChalmers1998}. A notebook, phone, or reminder system can be part of a person's practical cognitive process. A CDT that mediates memory, planning, therapy, communication, or decision-making may become more deeply integrated. Failure, manipulation, or epistemic displacement in such a system can therefore harm agency more profoundly than a mistaken recommendation from an ordinary external tool.

\subsection{Shadow Twins, Dual Use, and Manipulation}

A \term{shadow twin} is an individualized or proxy-based cognitive model constructed without the represented person's knowledge or meaningful consent. It may be built from behavioral traces, inferred traits, purchased data, workplace monitoring, educational records, platform interactions, medical data, or information supplied by others. Shadow twins are plausible because many CDT inputs already exist in institutional databases and platform logs.

Shadow twins and manipulation should be analyzed together because one of the main reasons to construct a shadow twin is to influence a person more effectively. Shadow twins can be used to predict vulnerability, rank trustworthiness, infer political receptivity, simulate likely reactions, personalize prices, test persuasion strategies, or select moments of intervention. They are ethically distinct from ordinary profiling because they claim to model cognition in a dynamic or operational way. The problem is not only privacy invasion. It is the creation of an actionable representation of the person outside the person's control. Surveillance capitalism, contextual privacy, and inferential privacy scholarship show how behavioral data can become prediction and influence infrastructure \cite{Zuboff2019,Nissenbaum2010Privacy,BarocasNissenbaum2014BigData,WachterMittelstadt2019ReasonableInferences}. Online manipulation scholarship has shown how informational environments can be structured to shape choices without reflective awareness \cite{Susser2019OnlineManipulation,Susser2019AdaptiveChoice,Calo2014DigitalMarketManipulation}. Generative models and automated influence operations intensify this concern because they can produce adaptive, scalable, and personalized persuasive content \cite{Matz2017PsychologicalTargeting,Matz2024PersonalizedPersuasion,HackenburgMargetts2024Microtargeting,Goldstein2023InfluenceOps}.

The simulated-patient case illustrates benign and harmful uses of the same capability. A clinic may simulate interventions to reduce harm. An insurer may simulate non-adherence to adjust coverage. An employer may simulate burnout or union interest. A political campaign may simulate persuasion pathways. A platform may simulate the moment when a user is most emotionally vulnerable to a message. CDT governance must therefore treat dual use as a default condition, not an exceptional misuse case. Technical safeguards inside user-facing products are insufficient if third parties can construct shadow twins from data ecosystems that already surround people.

This creates a manipulation risk that differs from generic recommendation. A recommender may optimize for engagement; a CDT may optimize through a model of deliberative vulnerability. A CDT may know which arguments a person accepts when tired, which relatives influence them, which fears are most motivating, and which interventions bypass reflection. The same prediction that helps a therapist avoid shame-inducing messages could help a platform select the most effective exploitative message. Intervention capacity should therefore be a governance trigger independent of accuracy. A highly accurate vulnerability model may be more dangerous, not less, when held by actors with incentives to manipulate.

\subsection{Simulation Without Participation}

CDTs can turn persons into sites of simulated experimentation. This is the central risk in the simulated-patient case. The clinic tests interventions on Diego's model, then acts on Diego. Such simulation may reduce harm, especially where real-world trial and error is costly or dangerous. But simulation can also substitute for participation. The patient becomes an object of prediction rather than a participant in interpretation.

This problem extends beyond healthcare. Schools may simulate how a student would respond to discipline. Employers may simulate whether an employee would resign, unionize, or burn out. Platforms may simulate how users would respond to interface changes. Policymakers may use simulated publics to forecast compliance, protest, or voting behavior. Social simulation can support research and policy analysis, but it can also become a substitute for consultation, consent, and democratic participation \cite{Aher2023TuringExperiment,Argyle2023OutOfOneMany,Park2024GenerativeSimulation,Mou2025SocialSimulationSurvey,Gao2024AgentBasedSimulation}.

The ethical problem is not simulation as such. It is representational authority. If a system claims that Diego would accept a treatment plan, or that a simulated community would accept a policy, decision-makers may treat the simulation as a proxy for the people themselves. This risk is amplified when simulated populations are built from data collected without meaningful consent (shadow twins), when marginalized groups are underrepresented or misrepresented, or when institutions prefer simulation because actual participation would be slower, more costly, or more resistant.

\subsection{Delegated Proxy Action and Source Gaps}

The productivity-twin case raises a different problem: a CDT may act as the person in social and institutional contexts. When Maya's twin sends a message, accepts a meeting, makes a promise, or gives feedback, others may interpret the action as Maya's own judgment. This blurs the line between action by a person, action for a person, and action about a person.

This problem is not simply a repeat of the accountability axis. Accountability asks who answers when harm occurs. Delegated proxy action asks what kind of social act occurred in the first place. A message signed by Maya, a message labeled as ``Maya's CDT,'' and a message drafted by a CDT but approved by Maya carry different meanings. They create different expectations of authenticity, reliance, and responsibility. The CDT-specific risk is that high-fidelity imitation makes proxy output socially legible as personal endorsement even when the represented person did not reflectively endorse it.

Delegated proxy action creates four gaps. First, there is a \term{source gap}: recipients may not know whether an action came from the person, the CDT, or a human using CDT assistance. Second, there is an \term{authorization gap}: the CDT may act within a broad permission category while exceeding what the person would have endorsed in context. Third, there is an \term{interpretation gap}: others may attribute the CDT's style, tone, or decision to the represented person. Fourth, there is a \term{liability gap}: providers, deployers, and users may each deny responsibility by pointing to the others. These gaps can create moral crumple zones, where humans absorb blame for failures of automated systems they did not meaningfully control \cite{Elish2019}. They also motivate authenticated delegation and auditable provenance for agents acting on behalf of people \cite{South2025AuthenticatedDelegation,Kroll2017AccountableAlgorithms}.

The solution is not to prohibit all delegation. Delegation is common and often valuable. The CDT-specific issue is that the delegated agent is socially framed as a cognitive representation of the person. A travel assistant that books a flight is not the same as an AI self that writes in Maya's style, reasons from her documents, and signs messages in her name. Higher-fidelity representation can increase usefulness while increasing the likelihood that others mistake proxy output for personal endorsement.

\begin{table*}[t]
\centering
\rowcolors{2}{lightgray}{white}
\small
\begin{tabular}{p{1.55in}p{3.0in}p{2.05in}}
\hline
\term{High-risk CDT requirement} & \term{Concrete implementation} & \term{Primary purpose} \\
\hline
Layered consent & Separate permissions for data collection, cognitive modeling, simulation, intervention, proxy action, third-party sharing, and secondary use & Prevent consent laundering \\
Purpose-binding enforcement & Data tags, use restrictions, access controls, downstream contracts, logs, and review for new uses & Prevent silent repurposing \\
Robustness and validity evaluation & Test construct validity, calibration, subgroup performance, temporal drift, distribution shift, missing data, feedback loops, and spurious proxies & Reduce non-robust cognitive claims \\
Audit logs & Record data provenance, model version, task instructions, uncertainty, permissions, human approvals, institutional uptake, and external actions & Enable investigation and redress \\
Proxy-action labels & Disclose when messages, commitments, decisions, or simulations are produced by or materially shaped by a CDT & Preserve source clarity and authenticity \\
Contestation procedure & Provide channels to dispute inferences, simulations, actions, permissions, and institutional uptake & Support rights and accountability \\
Revocation and model retirement & Permit pausing, narrowing, exporting, deactivation, and retirement of stale or unsafe models & Prevent frozen or unwanted representation \\
Independent review & Require risk-proportionate review for institutional simulations and high-stakes proxy action & Govern experimentation and dual use \\
Availability and refusal protections & Ensure access to socially important CDT benefits while preserving meaningful non-use & Prevent exclusion and coercion \\
Proxy-power constraints & Limit high-authority institutional scaling in contexts such as employment, education, politics, and public services & Prevent asymmetric proxy power \\
\hline
\end{tabular}
\caption{Concrete requirements for high-risk CDTs. These requirements are cumulative: high-risk systems need layered controls rather than a single transparency notice or consent form.}
\label{tab:highrisk}
\end{table*}

\section{Governance Gaps for Cognitive Digital Twins}

Existing governance frameworks partially address CDT risks, but they do so unevenly. When CDT harms look like privacy violations, automated decisions, manipulative systems, unsafe agents, or poorly validated models, existing tools can help. The gap is that CDTs combine these concerns in a distinctive governance object: a person-directed cognitive representation that may be simulated, trusted, delegated to, or used as a proxy for participation. This section asks what existing frameworks capture, what they miss, and why high-risk CDTs require governance at the level of cognitive representation itself. Along with other frameworks, we primarily reference the EU legislation as they have a well-developed legal framework. 

\subsection{Data Rights Do Not Exhaust Cognitive Control}

Data protection law is directly relevant to CDTs because many CDTs rely on personal data and, in clinical or mental-health contexts, special-category data (GDPR, Article 9(1)). The GDPR's principles of lawfulness, fairness, transparency, purpose limitation, data minimization, accuracy, storage limitation, integrity, confidentiality, and accountability (GDPR, Article 5(1)) are important CDT safeguards \cite{GDPR2016}. Its data-subject rights also matter: represented persons may have rights to access, rectification, erasure, restriction, portability, objection, and safeguards for automated decision-making (GDPR, Articles 12--22). These tools support CDT governance by giving people some control over data inputs and some ability to challenge outputs.

Yet data rights do not exhaust cognitive control. First, consent to data processing is not the same as consent to being cognitively modeled, simulated, or represented by a proxy. Diego might consent to clinical data use without understanding that the clinic will run counterfactual simulations of his likely reactions. Second, rights of access and correction may be difficult to exercise when the relevant representation is an inferred cognitive construct, a model state, an embedding, or a simulated response rather than an easily editable data field \cite{WachterMittelstadt2019ReasonableInferences}. Third, proxy and group-based CDTs may evade ordinary individual-rights framings when institutions claim they are modeling a class of people, even though the model is later applied as evidence about a specific person \cite{BarocasNissenbaum2014BigData,Nissenbaum2010Privacy}.

For CDTs, GDPR-style rights should therefore be extended into rights to know when one is modeled or simulated, to inspect high-level cognitive claims, to contest inferences, to restrict contexts of use, and to retire or freeze models where feasible. These extensions do not replace data protection law. They identify a CDT-specific control layer that data protection alone does not supply.

\subsection{Decision and Risk Categories Trigger Too Late}

Automated-decision and risk-based AI governance are directly relevant because CDT outputs may influence treatment, employment, education, finance, public services, policing, migration, or other high-stakes domains. The GDPR's Article 22 and related safeguards address some decisions based solely on automated processing that produce legal or similarly significant effects \cite{GDPR2016}. The EU AI Act imposes requirements on many high-risk systems, including risk management, data governance, technical documentation, record keeping, transparency, human oversight, accuracy, robustness, and cybersecurity, and it also prohibits certain manipulative or deceptive techniques, vulnerability exploitation, some social scoring, and some emotion-recognition uses \cite{EUAIAct2024}. These tools matter for CDTs, especially in institutional domains and for shadow twins used in workplace monitoring, political persuasion, or vulnerability targeting.

The gap is that CDT harms often arise before a final automated decision or outside existing risk categories. A therapist may use a CDT to interpret Diego as non-adherent before any formal decision is made. A manager may use a productivity twin's inferred communication style to decide how much autonomy to give Maya. A campaign may use a shadow twin to test persuasion strategies without producing a legally significant decision about the represented person. A clinical simulation CDT may influence treatment while remaining human-in-the-loop, and a simulated-publics tool may shape policy strategy without directly deciding about any individual. CDT governance therefore needs earlier triggers based on cognitive claim scope, intervention capacity, authority, autonomy, institutional uptake, and whether simulation substitutes for participation or contestation.

\subsection{Agent Governance Misses Non-Agentic Simulation}

Governance for autonomous agents usually focuses on external action: what tools the system can use, whether it can initiate tasks, what permissions it has, and how to prevent unsafe execution \cite{South2025AuthenticatedDelegation,Amodei2016ConcreteProblems}. These questions are essential for productivity twins like Maya's. A CDT that sends messages, schedules meetings, signs forms, negotiates services, or commits a person to deadlines requires clear authorization boundaries, source labels, logs, and revocation mechanisms.

But agent governance does not cover all CDT risks. Diego's clinical CDT may have no direct external authority. It may not send messages, spend money, or execute tasks. Yet it can still shape care because clinicians act on its simulations. A non-agentic CDT can therefore have high institutional influence even when it has low direct agency. CDT governance must therefore cover both proxy action and non-agentic simulation.

\subsection{Risk Management Needs Cognitive-Claim Discipline}

Risk-management frameworks such as the NIST AI RMF provide useful language for CDT governance. The RMF emphasizes validity, reliability, safety, security, resilience, accountability, transparency, explainability, privacy, and fairness through the functions Govern, Map, Measure, and Manage \cite{NIST2023AIRMF}. Model cards and datasheets also offer useful templates for documenting intended uses, evaluation results, dataset context, and limitations \cite{Mitchell2019ModelCards,Gebru2021Datasheets}. These tools are valuable for CDTs, especially where robustness, uncertainty, monitoring, and accountability are central.

However, generic risk management does not specify what counts as a valid cognitive claim. The problem is not only system accuracy, but also whether the system is measuring or simulating the cognitive phenomenon it claims to represent. CDT governance therefore requires cognitive-claim discipline: high-risk systems should document what cognitive constructs they claim to model, how those constructs are operationalized, what evidence supports them, what uncertainty remains, which uses are prohibited, and how represented persons can contest or contextualize the claims.

\subsection{Human-Subjects Ethics Does Not Clearly Cover Operational Simulation}

Simulation without participation creates a further gap. Human-subjects ethics offers useful principles for research involving people, including respect for persons, beneficence, and justice in the Belmont Report \cite{BelmontReport1979}. The Menlo Report extends related principles to information and communication technology research, including respect for law and public interest \cite{MenloReport2012}. These frameworks are relevant when people are modeled as targets of intervention.

However, many CDT simulations may not be classified as research. A clinic may simulate patient responses as part of treatment planning. An employer may simulate attrition or burnout as part of workforce management. A platform may simulate user reactions as part of product optimization. A policymaker may use simulated publics to anticipate compliance or opposition. These practices can be human-subjects-adjacent even when they are operational, commercial, or administrative rather than formally experimental. Institutions may rely on simulations of people while avoiding the duties that would attach to direct consultation, consent, or experimentation. For high-risk institutional simulation, CDT governance should therefore require review proportional to risk, and simulated persons and publics should supplement participation, not replace it.
\subsection{The Cross-Cutting Gap: Cognitive Representation}
Across these frameworks, the common gap is that existing governance often regulates data, decisions, or actions, while CDTs introduce risks at the level of cognitive representation. The critical event may be neither data collection nor final decision. It may be the moment a model's claim about a person's cognition becomes operational: a clinician trusts a simulated reaction, an employer trusts an inferred disposition, a campaign trusts a vulnerability model, or a recipient trusts a proxy message as personal endorsement. CDT governance must therefore add controls for modeling, simulation, interpretation, and proxy identity, not only controls for data processing, automated decision-making, and external action.

\section{Governance Recommendations}

The following recommendations are intended for systems that make person-directed cognitive claims, simulate cognitive responses, or act as cognitive proxies. Table \ref{tab:highrisk} summarizes the concrete requirements for high-risk CDTs. A CDT should be treated as high-risk when it is used in domains such as healthcare, education, employment, finance, criminal justice, border control, political persuasion, housing, or public-service access; when it is used by an institution about a person rather than by the represented person; when it has high authority or autonomy; when it targets vulnerable populations; when it can shape beliefs, emotions, choices, opportunities, or legal status; or when it is used without meaningful consent.

\subsection{Discipline Cognitive Representation}

CDTs should practice \term{evidentiary humility}. A system should not claim to model ``the mind'' without specifying what it models, from which data, at what level of uncertainty, for which population, and for which use. A system may have evidence that it predicts missed appointments, but not that it knows a person's motivation. It may estimate confusion in a tutoring task, but not general intelligence or moral character. It may imitate Maya's email style, but not know what Maya reflectively endorses.

Claim discipline requires a CDT system card documenting constructs, data sources, proxy validity, uncertainty, intended uses, prohibited uses, known failure modes, and subgroup performance. Model cards and datasheets provide useful templates for communicating intended use, evaluation, and dataset context \cite{Mitchell2019ModelCards,Gebru2021Datasheets}. CDT documentation should add cognitive-specific disclosures: which mental constructs are inferred, how they were operationalized, what evidence supports them, which contexts are out of scope, and how represented persons can contest them.

High-risk CDTs also require CDT-specific evaluation before deployment. Behavioral imitation is not enough. A CDT that predicts what a person will say or do may still misunderstand the person's reasons, values, constraints, or context. Evaluation should include construct validation, calibration, uncertainty reporting, subgroup and intersectional performance, temporal robustness, sensitivity to missing data, distribution-shift testing, feedback-loop analysis, red-teaming for manipulation and dual use, and tests of whether users and institutions over-trust the system \cite{AndrusGilbert2019Measurement,JacobsWallach2021Measurement,NIST2023AIRMF}. For simulation CDTs, evaluation should test counterfactual stability. For agentic CDTs, evaluation should test authorization boundaries, refusal behavior, and traceability. Interpretability is valuable because CDTs make claims about internal states and processes, but governance should not depend on solving interpretability completely \cite{DoshiVelezKim2017,Rudin2019,Bereska2024MechanisticInterpretability}.

\subsection{Preserve Consent, Purpose, and Control}

Dynamic consent is an existing approach developed in biomedical research to support ongoing, granular, digitally mediated consent preferences and participant engagement \cite{Kaye2015DynamicConsent,Williams2015DynamicConsent}. CDT governance should build on this literature, while adding consent layers for data collection, cognitive modeling, simulation, intervention, proxy action, third-party sharing, and secondary use. Diego may consent to treatment support but reject insurance access, family simulations, or employment screening. Maya may allow her twin to draft messages but not send them, attend meetings but not commit to deadlines, or use work documents but not private messages. Consent should be ongoing, contextual, revocable, and understandable. Emergency exceptions should be narrow, logged, justified, and reviewable.

Purpose binding restricts the CDT to agreed-upon uses and prevents silent repurposing. A mental-health CDT should not be used to evaluate employability. A memory-support CDT should not determine insurance premiums. A learning CDT should not become a disciplinary prediction tool without renewed authorization, independent scrutiny, and heightened safeguards. Purpose binding requires legal, organizational, and technical mechanisms: contracts and institutional policies should prohibit unauthorized secondary uses; review procedures should govern proposed new uses; and access controls, data tags, model-use restrictions, logging, permission scopes, and audit systems should make repurposing detectable.

Control also requires contestability, revocability, and model retirement. Represented persons and affected parties should be able to challenge CDT outputs, simulations, and actions without proving, in technical terms, that the cognitive model is false. They should be able to say: this does not represent me, this use was unauthorized, this inference is harmful, this action exceeded permission, this context was inappropriate, or this model is stale. Revocability means the ability to pause, narrow, export, or deactivate a CDT, subject to legitimate constraints such as safety, medical record requirements, or third-party rights. Model retirement is the stronger requirement that a CDT can be decommissioned when it is outdated, unsafe, unwanted, or no longer valid for its original purpose.

\subsection{Make Simulation and Proxy Action Traceable}

CDTs require traceability because their harms may arise through indirect institutional uptake or socially ambiguous proxy action. Any CDT that acts externally should make proxy status visible to appropriate parties. Messages, transactions, forms, and decisions should indicate whether they were produced by the person, the CDT, or a human acting with CDT assistance, subject to privacy and accessibility constraints. High-stakes actions should require confirmation thresholds and should record permissions, data sources, model version, task instructions, human confirmations, and escalation events \cite{South2025AuthenticatedDelegation,Raji2020AccountabilityGap,Kroll2017AccountableAlgorithms}.

Simulation CDTs also need traceability even when they do not act directly. In Diego's case, the clinic should be able to reconstruct whether a treatment change was influenced by the CDT, what data and model version were used, what uncertainty was reported, who reviewed the output, and whether Diego had an opportunity to contest or contextualize the inference. Logs should be protected, access-limited, and used for safety, accountability, contestation, and compliance, not generalized surveillance. Traceability should cover both \term{proxy action} and \term{institutional uptake}: the former asks whether an external act came from the person, the CDT, or a human using CDT assistance; the latter asks whether an organization relied on a CDT output when interpreting, classifying, intervening on, or deciding about a person.

\subsection{Govern Institutional Power}

Institutional simulation should be treated as human-subjects-adjacent even when it is not legally classified as research. The Belmont principles of respect for persons, beneficence, and justice remain relevant when people are modeled as targets of intervention \cite{BelmontReport1979}. The Menlo Report extends similar reasoning to information and communication technology research, including respect for law and public interest \cite{MenloReport2012}. High-risk simulations used for clinical treatment, employment, education, finance, policing, migration, public services, or political persuasion should require independent review proportional to risk. Review should examine consent layers, construct validity, data provenance, intended interventions, dual-use risks, disparate impact, subject participation, emergency exceptions, monitoring, and redress. Simulated publics should supplement, not replace, democratic participation.

CDT governance should also establish \term{availability floors} for socially important benefits and \term{proxy-power constraints} for high-authority institutional deployments. Availability floors may include public provision, insurance coverage, accessibility requirements, quality standards, and protections against inferior systems for marginalized groups. Proxy-power constraints may include limits on simultaneous high-authority agents, prohibitions on CDT-based political manipulation, restrictions on employment and education surveillance, special scrutiny for simulations of vulnerable populations, and requirements that humans remain reachable when rights or opportunities are at stake \cite{Eubanks2018AutomatingInequality,Kellogg2020AlgorithmsAtWork,RosenblatStark2016AlgorithmicLabor,Kulveit2025GradualDisempowerment}. The point is not to equalize all AI use. It is to prevent cognitive representation and delegated agency from becoming tools that only the powerful can use effectively, while everyone else becomes more predictable, targetable, and administratively exposed.

\section{Related Work}

Our argument connects several literatures around a single governance object: a person-directed cognitive representation that may be simulated, trusted, delegated to, or used as a proxy for participation. Digital twin scholarship provides the paradigm of dynamic representation, synchronization, prediction, and simulation \cite{Bruynseels2018}. Human and personal digital-twin work extends this paradigm to health and human-centered contexts, raising concerns about representation, consent, access, and control \cite{Braun2021,DeKerckhove2021PersonalDigitalTwin,Lin2024HDT,Katsoulakis2024DigitalTwinsHealth}. CDTs build on this paradigm but target cognition: self-understanding, deliberation, memory, communicative identity, and social agency.

Adjacent AI literatures show why CDTs are plausible and risky. Digital phenotyping and mental-health AI show how ordinary data streams can support mental-state inference \cite{Onnela2016DigitalPhenotyping,Torous2016DigitalPhenotyping,Mohr2017PersonalSensing,MartinezMartin2021DigitalPhenotyping}. Social simulation shows how language-model agents can approximate individual and collective behavior \cite{Park2023GenerativeAgents,Aher2023TuringExperiment,Argyle2023OutOfOneMany,Park2024GenerativeSimulation,Mou2025SocialSimulationSurvey,Gao2024AgentBasedSimulation}. Work on measurement validity, feedback loops, and system-wide fairness explains why CDT failures may emerge from constructs, deployment contexts, and institutional feedback, not only model error \cite{AndrusGilbert2019Measurement,JacobsWallach2021Measurement,Jiang2019FeedbackLoops,CruzCortesGhosh2020SystemWideFairness}.

Normative and governance literatures specify the risks. Neuroethics provides concepts such as cognitive liberty, mental privacy, mental integrity, and psychological continuity, while cognitive-biometrics work extends these concerns beyond neural data \cite{Ienca2017Neurorights,Ligthart2023MindingRights,Magee2024CognitiveBiometrics}. Work on AI extenders, cognitive offloading, delegation, adaptive choice architectures, epistemic injustice, AI safety, authenticated delegation, and interpretability frames CDT risks around cognitive extension, proxy authority, manipulation, epistemic displacement, agentic action, and monitoring \cite{HernandezVold2019AIExtenders,RiskoGilbert2016CognitiveOffloading,BaumerMcGee2019Speaking,Susser2019AdaptiveChoice,Kay2024EpistemicInjustice,Amodei2016ConcreteProblems,Carlsmith2022PowerSeeking,Krakovna2023PowerSeeking,South2025AuthenticatedDelegation,DoshiVelezKim2017,Rudin2019,Bereska2024MechanisticInterpretability}. Existing governance frameworks provide tools such as risk management, human oversight, automated-decision rights, purpose limitation, and restrictions on manipulation \cite{GDPR2016,EUAIAct2024,NIST2023AIRMF}. Our claim is that CDTs recombine these concerns around a cognitive representation that may be used to simulate, influence, classify, decide about, or act for a person.

\section{Conclusion and Open Questions}

Cognitive digital twins shift AI governance from systems that merely serve people to systems that represent people as cognitive objects. They may support memory, accessibility, care, education, research, creativity, and self-understanding. They may also enable frozen misrepresentation, non-robust inference, shadow twinning, simulation without participation, adaptive manipulation, posthumous exploitation, accountability gaps, and new forms of proxy power.

We proposed a working definition of CDTs, introduced the 5A framework, identified CDT-specific risks, analyzed governance gaps, and proposed concrete requirements for high-risk deployments. The core lesson is that CDT governance must treat cognition as ethically significant without mystifying it. CDTs do not need to be conscious to matter. They matter because people and institutions may rely on them as if they reveal what a person thinks, wants, would choose, or should become.

Several questions remain open. What evidentiary threshold should a system meet before claiming to model cognition? Which CDT uses should be categorically prohibited? How should law distinguish prediction, support, persuasion, manipulation, and simulated experimentation? What forms of consent and refusal are meaningful when models are continuously updated or institutionally constructed? How should proxy CDTs be governed when they are used as substitutes for specific persons? Can simulated publics ever inform democratic decision-making without replacing participation? How should posthumous and incapacity cases be governed? And how can societies preserve the benefits of cognitive assistance while preventing cognitive representation from becoming an infrastructure of control?

\bibliography{references}
\end{document}